\documentclass[conference]{IEEEtran}
\IEEEoverridecommandlockouts
\IEEEpubid{\makebox[\columnwidth]{978-1-6654-3274-0/21/\$31.00~\copyright2021 IEEE \hfill} \hspace{\columnsep}\makebox[\columnwidth]{ }}
\usepackage{cite}
\usepackage{microtype}
\usepackage{graphicx}
\usepackage{subfigure}
\usepackage{booktabs} 
\usepackage{wrapfig}
\usepackage{hyperref}

\usepackage[utf8]{inputenc} 
\usepackage[T1]{fontenc}    
\usepackage{hyperref}       
\usepackage{url}            
\usepackage{booktabs}       
\usepackage{nicefrac}       
\usepackage{microtype}      

\usepackage{tikz}
\usepackage{blindtext}
\usepackage{hyperref}
\usepackage{url}
\usepackage{float}
\usepackage{multirow}
\usepackage{booktabs}
\usepackage{caption}
\usepackage{comment}
\usepackage[titlenumbered,ruled]{algorithm2e}
\usepackage{algorithmicx}

\usepackage{array}
\usepackage{wrapfig}
\usepackage{fdsymbol}

\usepackage{bbm}
\usepackage{amsfonts}
\usepackage{amsmath}
\usepackage{graphicx}
\usepackage{textcomp}
\usepackage{xcolor}

\makeatletter
\newcommand{\removelatexerror}{\let\@latex@error\@gobble}
\makeatother

\def\BibTeX{{\rm B\kern-.05em{\sc i\kern-.025em b}\kern-.08em
    T\kern-.1667em\lower.7ex\hbox{E}\kern-.125emX}}
\begin{document}
\bstctlcite{IEEEexample:BSTcontrol}

\title{InstantNet: Automated Generation and Deployment of Instantaneously Switchable-Precision Networks
}

\makeatletter
\newcommand{\linebreakand}{%
  \end{@IEEEauthorhalign}
  \hfill\mbox{}\par
  \mbox{}\hfill\begin{@IEEEauthorhalign}
}
\makeatother

\author{\IEEEauthorblockN{Yonggan Fu}
\IEEEauthorblockA{\textit{Rice University} \\
yf22@rice.edu} 
\and
\IEEEauthorblockN{Zhongzhi Yu}
\IEEEauthorblockA{\textit{Rice University} \\
zy42@rice.edu}
\and
\IEEEauthorblockN{Yongan Zhang}
\IEEEauthorblockA{\textit{Rice University} \\
yz87@rice.edu}
\and
\IEEEauthorblockN{Yifan Jiang}
\IEEEauthorblockA{\textit{University of Texas at Austin} \\
yifanjiang97@utexas.edu}
\and
\IEEEauthorblockN{Chaojian Li}
\IEEEauthorblockA{\textit{Rice University} \\
cl114@rice.edu}
\linebreakand
\IEEEauthorblockN{Yongyuan Liang}
\IEEEauthorblockA{\textit{Sun Yat-sen University} \\
liangyy58@mail2.sysu.edu.cn}
\and
\IEEEauthorblockN{Mingchao Jiang}
\IEEEauthorblockA{\textit{Rice University} \\
mj33@rice.edu}
\and
\IEEEauthorblockN{Zhangyang Wang}
\IEEEauthorblockA{\textit{University of Texas at Austin} \\
atlaswang@utexas.edu}
\and
\IEEEauthorblockN{Yingyan Lin}
\IEEEauthorblockA{\textit{Rice University} \\
yingyan.lin@rice.edu}
}

\maketitle

\begin{abstract}

The promise of Deep Neural Network (DNN) powered Internet of Thing (IoT) devices has motivated a tremendous demand for automated solutions to enable fast development and deployment of efficient (1) DNNs equipped with instantaneous accuracy-efficiency trade-off capability to accommodate the time-varying resources at IoT devices and (2) dataflows to optimize DNNs' execution efficiency on different devices. Therefore, we propose InstantNet to automatically generate and deploy instantaneously switchable-precision networks which operate at variable bit-widths. Extensive experiments show that the proposed InstantNet consistently outperforms state-of-the-art designs. Our codes are available at: \underline{\href{https://github.com/RICE-EIC/InstantNet}{https://github.com/RICE-EIC/InstantNet}}.

\end{abstract}


 \vspace{-0.2cm}
\section{Introduction}

Powerful deep neural networks (DNNs)' prohibitive complexity calls for hardware efficient DNN solutions \cite{eyeriss,10.1145/3210240.3210337,8050797}. When it comes to DNNs' hardware efficiency in IoT devices, the model complexity (e.g., bit-widths), dataflows, and hardware architectures are major performance determinators.
Early works mostly provide \textit{static} solutions, i.e., once developed, the algorithm/dataflow/hardware are fixed, whereas IoT applications often have dynamic time/energy constraints over time. Recognizing this gap, recent works \cite{jin2019adabits,guerra2020switchable} have attempted to develop efficient DNNs with instantaneous accuracy-cost trade-off capability. For example, switchable-precision networks (SP-Nets) \cite{jin2019adabits,guerra2020switchable} can maintain a competitive accuracy under different bit-widths without fine-tuning under each bit-width, making it possible to allocate bit-widths on the fly for adapting IoT devices' instant resources over time. 

Despite SP-Nets' great promise \cite{jin2019adabits,guerra2020switchable}, there are still major challenges in enabling their deployment into numerous IoT devices. First, existing SP-Nets are manually designed, largely limiting their extensive adoption as \textit{each application} would require a \textit{different} SP-Net. 
Second, while the best dataflow for SP-Nets under \textit{different bit-widths} can be different and is an important determinator for their on-device efficiency \cite{venkatesanmagnet}, there is still a lack of a generic and publicly available framework that can be used to suggest optimal dataflows for SP-Nets under \textit{each of their bit-widths} on \textit{different IoT devices}. Both of the aforementioned hinder the fast development and deployment of SP-Nets powered DNN solutions for  
\textit{diverse} hardware platforms of IoT devices.   
To tackle the aforementioned challenges, we make the following contributions: 

\vspace{-0.1cm}



\begin{itemize}
    \item We propose InstantNet, an end-to-end framework that automates the development (i.e., the generation of SP-Nets given a dataset and target accuracy) and deployment (i.e., the generation of the optimal dataflows) of SP-Nets. To our best knowledge, InstantNet is \textbf{the first} to simultaneously target both development and deployment of SP-Nets.


    
    
    \item We develop switchable-precision neural architecture search (SP-NAS) that integrates an novel cascade distillation training to ensure that the generated 
    SP-Nets under all bit-widths achieve the same or better accuracy 
than both \textit{NAS generated} DNNs optimized for individual bit-widths and SOTA \textit{expert-designed} SP-Nets.
    
    \item We propose AutoMapper, which integrates a generic dataflow space and an evolutionary algorithm to navigate over the discrete and large mapping-method space and automatically search for optimal dataflows given a DNN (e.g., SP-Nets under a selected bit-width) and target device. 
    
    
    \item 
    Extensive experiments based on real-device measurements and hardware synthesis validate InstantNet's effectiveness in consistently outperforming
    SOTA designs, e.g., achieving 84.68\%  real-device Energy-Delay-Product improvement while boosting the accuracy by 1.44\%, over the most competitive competitor under the same settings.

\end{itemize}
\vspace{-0.3em}
\section{Related works}  
\textbf{Static and switchable-precision DNNs.}
DNN quantization aims to compress DNNs at the most fine-grained bit-level~\cite{fractrain, fu2021cpt}. 
To accommodate constrained and time-varying resources on IoT devices, 
SP-Nets~\cite{jin2019adabits, guerra2020switchable} aim for instantaneously switchable accuracy-efficiency trade-offs at the bit-level. 
However, designing such DNNs and the corresponding mapping methods for every scenario can be engineering-expensive and time consuming, considering the ever-increasing IoT devices with diverse hardware platforms and application requirements. As such, techniques that enable 
fast development and deployment of SP-Nets are highly desirable for expediting the deployment of affordable DNNs into numerous IoT devices.

\textbf{Neural Architecture Search for efficient DNNs.} 
To release human efforts from laborious manual design, NAS~\cite{zoph2016neural, fu2020autogandistiller}
have been introduced to enable the automatic search for efficient DNNs with both competitive accuracy and hardware efficiency given the datasets. 
\cite{wang2019haq, chen2018joint, wu2018mixed} incorporate quantization bit-widths into their search space and search for mixed-precision networks. However, all these NAS methods search for quantized DNNs with only one \textit{fixed} bit-width, lacking the capability to instantly adapt to other bit-widths without fine-tuning.

\textbf{Mapping DNNs to devices/hardware.} 
When deploying DNNs into IoT devices with diverse hardware architectures, one major factor that determines hardware efficiency is the dataflow
\cite{venkatesanmagnet}. For devices with application-specific integrated circuit (ASIC) or FPGA hardware,
various innovative dataflows \cite{eyeriss, Optimize_fpga_for_DNN, zhang2018dnnbuilder,10.1109/ISCA45697.2020.00082} have been developed to maximize the reuse opportunities. Recently, MAGNet has been proposed to automatically identify optimal dataflows and design parameters of a tiled architecture. However, its highly template-based design space, e.g., a pre-defined set of nested loop-orders, can restrict the generality and result in sub-optimal  performance. 
Despite its promising performance, 
the exploration to automatically identify optimal mapping methods for DNNs with different bit-widths has not yet been considered.

\begin{figure}[!t]
    \centering
    \includegraphics[width=0.48\textwidth]{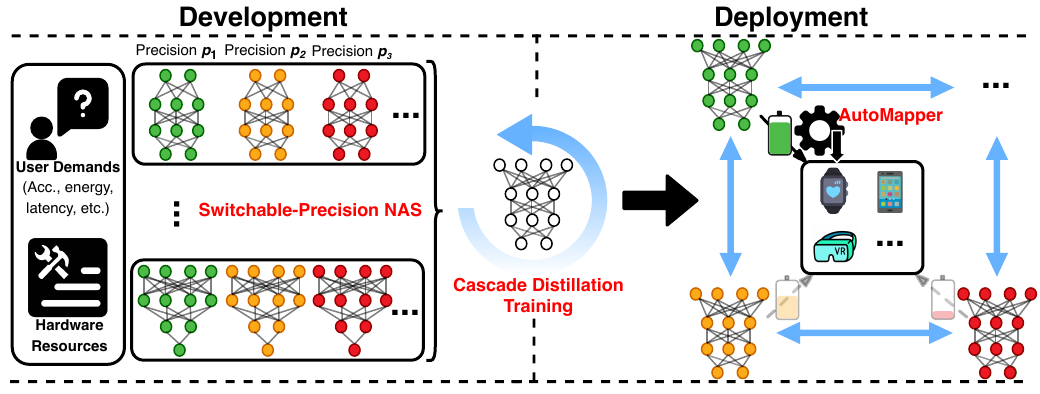}
    \caption{Overview of InstantNet, which 
    first generates SP-Nets with high accuracy under all bit-widths, and then suggests dataflows to maximize the generated SP-Nets' execution efficiency under different bit-widths on the target device. 
    }
    \label{fig:overview}
     \vspace{-1em}
\end{figure}

\vspace{-0.5em}
\section{The proposed InstantNet framework}
 \vspace{-0.1cm}
Here we present our InstantNet framework, starting from an overview and then its key enablers including cascade distillation training (CDT), SP-NAS, and AutoMapper.


\vspace{-0.1cm}
\subsection{InstantNet overview}
\label{sec:overview}
Fig.~\ref{fig:overview} shows an overview of InstantNet. Specifically, given the target application and device, it automates the development and deployment of SP-Nets. Specifically, InstantNet integrates two key enablers: (1) SP-NAS and (2) AutoMapper. SP-NAS incorporates an innovative cascade distillation to search for SP-Nets, providing IoT devices' desired instantaneous accuracy-efficiency trade-off capability. AutoMapper adopts a generic dataflow design space and an evolution-based algorithm to automatically search for optimal dataflows of SP-Nets under different bit-widths on the target device.

\begin{figure*}[!bt]
    \centering
    \includegraphics[width=0.8\textwidth]{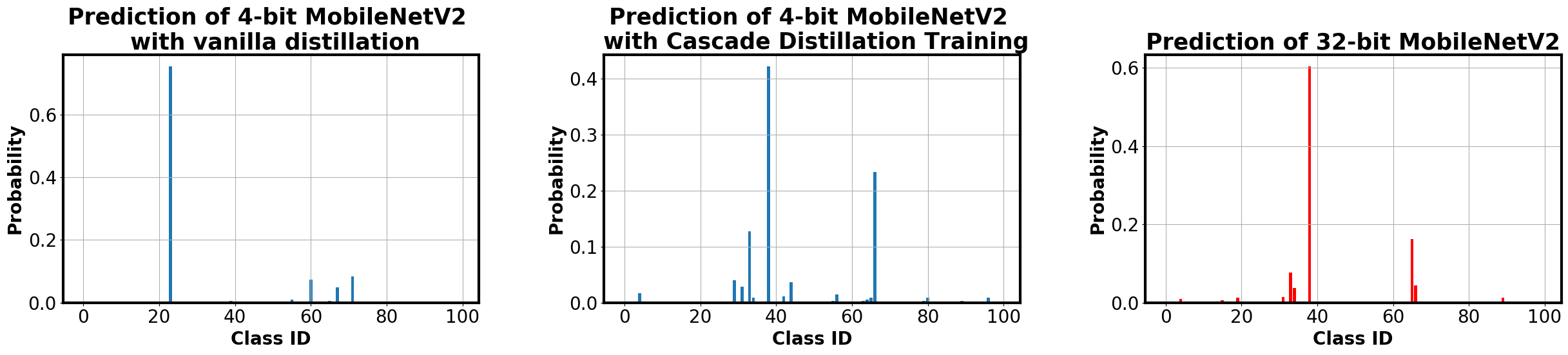}
    \vspace{-0.6em}
    \caption{Visualizing the prediction distribution of MobileNetV2 on CIFAR-100 under \textbf{(left)}: 4-bit training with vanilla distillation, \textbf{(middle)} 4-bit training with the proposed CDT, and \textbf{(right)} 32-bit training.}
    \label{fig:output}
    \vspace{-2em}
\end{figure*}

\vspace{-0.1cm}
\subsection{InstantNet training: Bit-Wise Cascade Distillation}
\label{sec:cdt}
\vspace{-0.1cm}
Unlike generic quantized DNNs optimized to maximize accuracy under one individual bit-width, InstantNet aims to generate SP-Nets of which the accuracy \textit{under all bit-widths} are the same or even higher than that of DNNs customized for individual bit-widths. The key challenge is to ensure high accuracy for lower bit-widths, which is particularly difficult for compact DNN models whose accuracy is more sensitive to quantization. For example, SOTA SP-Nets  ~\cite{guerra2020switchable} fails to work on lower bit-widths when being applied to MobileNetV2~\cite{sandler2018mobilenetv2}. 
The above challenge has motivated InstantNet's CDT method, which takes advantage of the fact that the quantization noises of SP-Nets under adjacent or closer bit-widths are smaller. Our hypothesis is that distillation between 
adjacent and closer bit-widths will help to more smoothly enforce the accuracy (or activation distribution) of SP-Nets under low bit-widths to approach their full-precision counterparts. In this way, CDT can simultaneously boost accuracy of SP-Nets under all bit-widths by enforcing SP-Nets under each bit-width to have distillation from \textbf{all higher bit-widths}: 

\vspace{-1em}
\begin{equation}
\vspace{-0.3em}
    \begin{split}
    L_{total} &= \frac{1}{N} \sum_{i=0}^{N-1} L_{train}^{cas}(Q_i(\omega))  \\
   where \; L_{train}^{cas}&(Q_i(\omega)) = L_{ce}(Q_i(\omega), label) \\
   + \beta \sum_{j=i+1}^{N-1}& L_{mse}(Q_i(\omega), SG(Q_j(\omega)))
    \end{split} \label{eqn:cdt}
\end{equation}

\noindent where $L_{total}$ is SP-Nets' average loss under all the $N$ candidate bit-widths, $L_{ce}$ and $L_{mse}$ are the cross-entropy and mean square error losses, respectively, $Q_i(\omega)$ is the SP-Net characterized with weights $\omega$ under the $i$-th bit-width,
$\beta$ is a trade-off parameter, and $SG$ is the stopping gradient function, i.e., gradient backpropagation from higher bit-widths is prohibited when calculating the distillation loss~\cite{guerra2020switchable}.

To verify the effectiveness of CDT, we visualize the prediction distribution (classification probability after softmax) of MobileNetV2 on CIFAR-100 under the bit-width set of 4, 8, 12, 16, 32 (quantized by SBM~\cite{banner2018scalable}) trained using different strategies in Fig.~\ref{fig:output}. 
We show the prediction distribution of the following three cases using a random sampled image from the test dataset to verify and visualize the effectiveness of our CDT: (1) 4-bit trained using vanilla distillation, i.e., only consider the distillation with 32-bit width, (2) 4-bit trained using our CDT technique and (3) the 32-bit trained network. We can observe that vanilla distillation fails to narrow the gap between 32-bit and the lowest 4-bit due to the large quantization error gap. This is actually a common phenomenon among efficient models with depthwise layers which are sensitive to low precision on all the considered test datasets, e.g., we observe that the validation accuracy of the 4-bit network with only the aforementioned vanilla distillation is around 1\%, indicating the failure of vanilla distillation for tackling the bit-width set with a large dynamic range. In contrast, our CDT notably helps the prediction distribution of the 4-bit network smoothly evolve to that of the 32-bit one, and also boost its accuracy to 71.21\%, verifying CDT's effectiveness.

\vspace{-0.1cm}
\subsection{InstantNet search: Switchable-Precision NAS }
\label{sec:banas}
\vspace{-0.1cm}
Here we introduce another key enabler of InstantNet, SP-NAS. To our best knowledge, InstantNet is \textbf{the first} to address \textit{how to automatically generate networks which naturally favor working under various bit-widths}. In addition, to resolve the performance bottleneck in SOTA SP-Nets (manually designed) ~\cite{jin2019adabits, guerra2020switchable}, i.e., large accuracy degradation under the lowest bit-width, we develop a heterogeneous scheme for updating the weights and architecture parameters. Specifically, we update the weights based on our CDT method (see Eq.~\ref{eqn:cdt}) which explicitly incorporates switchable-bit property into the training process; and for updating the architecture parameters of SP-Net, we adopt \textit{only the weights under the lowest bit-width}, for generating networks forced to inherently tackle SP-Nets' bottleneck of high accuracy loss under the lowest bit-width:  

\vspace{-1em}
\begin{equation} \label{eqn:banas}
\vspace{-0.3em}
    \begin{split}
    & \min \limits_{\alpha} L_{val}(Q_0(\omega^*), \alpha)+\lambda L_{eff}(\alpha) \\
    & s.t. \quad \omega^* = \underset{\omega}{\arg\min} \,\,  \frac{1}{N} \sum_{i=0}^{N-1} L_{train}^{cas}(Q_i(\omega),  \alpha)
    \end{split}
\end{equation}

\noindent where $\omega$ and $ \alpha$ are the supernet's weights \cite{liu2018darts} and architecture parameters, respectively, 
$L_{eff}$ is an efficiency loss (e.g., energy cost), and $Q_0(\omega)$ is the SP-Net under the lowest bit-width. 
Without loss of generality, here we adopt SOTA differentiable NAS~\cite{liu2018darts} and search space~\cite{wu2019fbnet}.

\begin{figure*}[!t]
    \centering
    \includegraphics[width=0.8\textwidth]{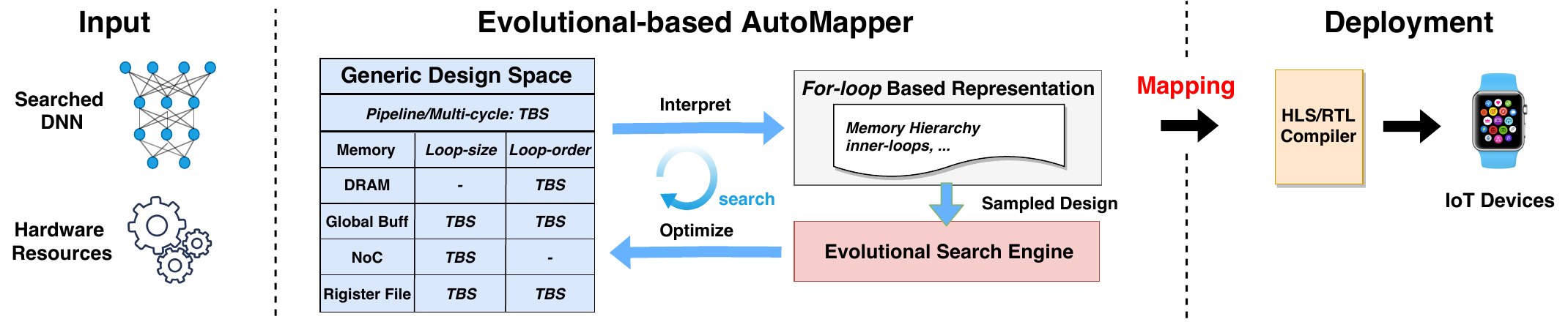}
    \vspace{-0.5em}
    \caption{Overview of the goal, generic dataflow space, and InstantNet's AutoMapper, where TBS denotes ``to be searched''.}
    \label{fig:map_overview}
    \vspace{-1em}
\end{figure*}

\vspace{-0.1cm}
\subsection{InstantNet deploy: Evolution-based AutoMapper}
\vspace{-0.1cm}
This subsection introduces InstantNet's AutoMapper, of which an overview is shown in Fig.~\ref{fig:map_overview}. Motivated by the fact that different mapping methods can have orders-of-magnitude difference in hardware efficiency \cite{venkatesanmagnet}, AutoMapper aims to accept (1) DNNs (e.g., SP-Nets generated by our SP-NAS), (2) the target device, and (3) target hardware efficiency, and then generate mapping methods that maximize both the task accuracy and hardware efficiency of the given SP-Nets under all bit-widths when being executed on the target device. 



\textbf{Generic Dataflow Design Space.}
A generic dataflow design space is a prerequisite for effective algorithmic exploration and optimization of on-device dataflows, yet is challenging to develop. There are numerous choices for how to temporally and spatially schedule all the DNN's operations to be executed in the target accelerators. Specifically,
as there are many more operations in DNNs than the number of operations (e.g., $19.6E+9$ \cite{simonyan2014very} vs. 900 MACs \cite{xilinxzc706} assuming a 16-bit precision) an IoT device can execute in each clock cycle, numerous possible dataflows exist for running DNNs on a device. 
To tackle the aforementioned challenge, we propose a generic design space for on-device dataflows, which (1) covers all design choices for generalization and (2) is easy to understand for ease of adoption. Our proposed space leverages commonly used nested \textit{for-loop} descriptions~\cite{eyeriss,DNNCHIPPREDICTOR}. For better illustration, here we describe the high-level principles. 
From a nested \textit{for-loop} description, our dataflow space extracts all possible choices characterized by the following factors:

\textit{loop-order}: the processing order of each dimension within each memory hierarchy, 
and can be derived from all possible permuted choices without overlap.


\textit{loop-size}: the no. of operations in one iteration of a specific dimension, which can not be easily determined. We design a simple analytical algorithm to derive all possible choices. 

\textit{Pipeline/multi-cycle:} use pipeline or multi-cycle. The former processes a small chunk of each layer in a pipeline manner, while the latter processes all the layers sequentially. 

Considering AlexNet \cite{krizhevsky2012imagenet} and six layers of nested loops, there are over \textbf{$10^{27}$ total number of discrete mapping-method choices}, posing a great need for developing efficient and effective search algorithms.

\begin{figure}[!b]
    \vspace{-1em}
    \begin{minipage}{\linewidth}
    \removelatexerror
    \begin{algorithm}[H]
    \DontPrintSemicolon
    \KwIn{Efficiency Goal, DNN, Design Space (DS)}
    \KwOut{Optimal algorithm-to-device mapping}

      Build a $pool$ with $n$ random samples from DS
      
      \While{\textit{Efficiency Goal not met}}
      {
        \uIf{$size(pool) \leq n$}
        {
          \For{$m$ iterations}
          {

            Random Pick  $p  \in pool$\;
            Random Perturb $k$ features of $p$

          Add $p$ to $pool$
          
         }
        }
        \Else
        {
            Rank the samples in $pool$\ with the given DNN
            
            Remove the worst $m$ samples from  \textit{pool}\; 
        }
      }
      \Return{optimal mapping in $pool$}
    \caption{Evolutionary AutoMapper}\label{alg:eaalg}
    \end{algorithm}
\end{minipage}
\end{figure}

\textbf{Evolutionary Search Algorithm.} To navigate the large and discrete space of mapping methods, we adopt an evolutionary based search algorithm, considering that evolutionary algorithms have more exploitation than random search and are better suited for the highly discrete space \cite{google_ev,genesys}.
Specifically, we will keep track of the hardware efficiency ranking of the current sampled mapping methods at each iteration. Afterwards, if the pool size of current samples is smaller than a specified value, we select a few of the best performing sampled mapping methods and randomly perturb a small number of their features associated with the aforementioned design factors to generate new mapping methods to be evaluated in the next iteration; otherwise, new mapping methods with completely randomly selected design factors will be generated.
We summarize our Evolutionary AutoMapper in Alg.\ref{alg:eaalg}. 


\begin{table}[!t]
  \centering
  \caption{InstantNet's CDT over SBM~\cite{banner2018scalable} (SOTA training for quantized DNNs) and SOTA SP-Nets (SP~\cite{guerra2020switchable} and AdaBits~\cite{jin2019adabits}) on \textbf{MobileNetV2} and CIFAR-100 in terms of test accuracy (\%), where the values in the bracket represent the accuracy drop of the baseline methods compared to our CDT.
}
\resizebox{0.95\linewidth}{!}{
    \begin{tabular}{ccccc}
    \toprule
    \multicolumn{1}{c}{Bit-widths} & \multicolumn{1}{c}{SBM~\cite{banner2018scalable}} & \multicolumn{1}{c}{SP~\cite{guerra2020switchable}} & \multicolumn{1}{c}{AdaBits~\cite{jin2019adabits}} & CDT (Proposed) \\
    \midrule
    4      & 70.55 \textbf{(-0.60)}  & 66.75 \textbf{(-4.40)}  & 68.07 \textbf{(-3.08)}  & \textbf{71.15}\\
    8      & 74.40 \textbf{(-0.72)}  & 71.69 \textbf{(-3.43)} & 73.86 \textbf{(-1.26)}  & \textbf{75.12} \\
    12     & 74.87 \textbf{(-0.16)} & 74.16 \textbf{(-0.87)} & 73.65 \textbf{(-1.38)}  & \textbf{75.03} \\
    16     & 75.03 \textbf{(-0.19)} & 74.23 \textbf{(-0.99)}  & 73.87 \textbf{(-1.35)}  & \textbf{75.22} \\
    32     & 75.23 \textbf{(+0.25)} & 74.11 \textbf{(-0.87)}  & 74.51 \textbf{(-0.47)}  & \textbf{74.98}  \\
    \midrule
    \midrule
    4      & 70.55 \textbf{(-0.53)}  & 67.63 \textbf{(-3.45)}  & 68.37 \textbf{(-2.71)}  & \textbf{71.08} \\
    5      & 74.13 \textbf{(-0.32)} & 72.95 \textbf{(-1.50)}  & 73.52 \textbf{(-0.93)}  & \textbf{74.45} \\
    6      & 74.69 \textbf{(-0.33)} & 74.15 \textbf{(-0.87)}  & 74.60 \textbf{(-0.42)}   & \textbf{75.02} \\
    8      & 74.40 \textbf{(-0.64)} & 74.99 \textbf{(-0.05)}  & 75.02 \textbf{(-0.02)}  & \textbf{75.04} \\
    \bottomrule
    \end{tabular}%
}
 \vspace{-1.2em}
  \label{tab:cascade}%
\end{table}%

\vspace{-0.1cm}
\section{Experiment results}
We first describe our experiment setup and then evaluate each enabler of InstantNet, i.e., CDT, SP-NAS, and AutoMapper. After that, we benchmark InstantNet over SOTA SP-Nets on SOTA accelerators \cite{zhang2018dnnbuilder, XilinxCH65, eyeriss}. 

\vspace{-0.1cm}
\subsection{Experiment setup}
\label{sec:exp_setup}
\vspace{-0.1cm}
\subsubsection{Algorithm experiment setup}
\textbf{Datasets \& Baselines.} We consider three datasets (CIFAR-10/CIFAR-100/ImageNet), and evaluate InstantNet over (1) all currently published SP-Nets (AdaBits~\cite{jin2019adabits} and SP~\cite{guerra2020switchable}) with the DoReFa~\cite{zhou2016dorefa} quantizer and (2) a SOTA quantized DNN method SBM~\cite{banner2018scalable} to train a SOTA compact DNN MobileNetV2~\cite{sandler2018mobilenetv2} under individual bits.

\textbf{Search and training on CIFAR-10/100 and ImageNet.}  \underline{Search space:} we adopt the same search space as~\cite{wu2019fbnet} except the stride settings for each group to adapt to the resolution of the input images in CIFAR-10/100.
\underline{Search settings.} On CIFAR-10/100, we search for 50 epochs with batch size 64. In particular, we (1) update the supernet weights with our cascade distillation technique as in Eq.(2) on half of the training dataset using an SGD optimizer with a momentum of 0.9 and an initial learning rate (LR)  0.025 at a cosine decay, and (2) update network architecture parameters with the lowest bit-width as in Eq.(2) on the other half of the training dataset using an Adam optimizer with a momentum of 0.9 and a fixed LR 3e-4. We apply gumbel softmax on the architecture parameters as the contributing coefficients of each option to the supernet (following~\cite{wu2019fbnet}), where the initial temperature is 3 and  then decayed by 0.94 at each epoch. On ImageNet, we follow the same hyper-parameter settings for the network search as~\cite{wu2019fbnet}.
\underline{Evaluate the derived networks:} for training the derived networks from scratch using our CDT, on CIFAR-10/100 we adopt an SGD optimizer with a momentum of 0.9 and an initial LR 0.025 at a cosine decay. Each network is trained for 200 epochs with batch size 128. On ImageNet, we follow~\cite{wu2019fbnet}.

\subsubsection{Hardware experiment setup}
\textbf{Implementation methodology.} We consider two commonly used IoT hardware platforms, i.e., ASIC and FPGA, for evaluating our AutoMapper. Specifically, for FPGA, we adopt the Vivado HLx design tool-flow where we first synthesize the mapping-method design in C++ via Vivado HLS, and then plug the HLS exported IPs into a Vivado IP integrator to generate the corresponding bit streams, which are programmed into the FPGA board for on-board 
execution and measurements; for ASIC, we synthesize the Verilog designs based on the generated dataflows using a Synopsys Design Compiler on a commercial CMOS technology, and then place and route using a Synopsys IC Compiler for obtaining the resulting design's actual area.

\textbf{Baselines.}
We evaluate AutoMapper over expert/tool generated SOTA dataflows for both FPGA and ASIC platforms, including DNNBuilder~\cite{zhang2018dnnbuilder} and CHaiDNN~\cite{XilinxCH65} for FPGA, and Eyeriss~\cite{eyeriss} and MAGNet~\cite{venkatesanmagnet} for ASIC. For DNNBuilder~\cite{zhang2018dnnbuilder}, MAGNet~\cite{venkatesanmagnet} and CHaiDNN~\cite{XilinxCH65}, we use their reported results; For Eyeriss~\cite{eyeriss}, we use their own published and verified simulator~\cite{Gao2017Tetris} to obtain their results.

\vspace{-0.1cm}
\subsection{Ablation study of InstantNet: CDT}
\label{sec:exp_cd}
\vspace{-0.1cm}
\textbf{Experiment settings.} For evaluating InstantNet's CDT, we benchmark it over an SOTA quantized DNN training method (independently train DNNs at each bit-width) and two SP-Nets (AdaBits~\cite{jin2019adabits} and SP~\cite{guerra2020switchable}). In light of our IoT application goal, we consider MobileNetV2~\cite{sandler2018mobilenetv2} (an SOTA efficient model balancing task accuracy and hardware efficiency) with CIFAR-100, and adopt two different bit-width sets with both large and narrow bit-width dynamic ranges. Without losing generality, our CDT is designed with SOTA quantizer SBM~\cite{banner2018scalable} and switchable batch normalization as in SP~\cite{guerra2020switchable}.

\textbf{Results and analysis.}
From Tab.~\ref{tab:cascade}, we have three observations: (1) our CDT consistently outperforms the two SP-Net baselines under all the bit-widths, verifying CDT's effectiveness and our hypothesis that progressively distilling from all higher bit-widths can help more smoothly approach accuracy of the full-precision; (2) CDT is particularly capable of boosting accuracy in low bit-widths which has been shown to be the bottleneck in exiting SP-Nets~\cite{jin2019adabits}, e.g., a 2.71\%$\sim$4.4\% higher accuracy on the lowest 4-bit over the two SP-Net baselines; and (3) CDT always achieves a higher or comparable accuracy over the SOTA quantized DNN training method SBM that independently trains and optimizes each individual bit-width: for bit-widths ranging from 4-bit to 8-bit, CDT achieves 0.32\%$\sim$0.72\% improvement in accuracy over SBM, indicating the effectiveness of our CDT in boosting DNNs' accuracies under lower bit-widths.

\begin{table}[!t]
  \centering
  \caption{CDT over independently trained SBM~\cite{banner2018scalable} on \textbf{ResNet-38}, where the values in the bracket represent CDT's accuracy gain over SBM (the higher, the better)}. 
  \resizebox{0.8\linewidth}{!}{
    \begin{tabular}{ccc|cc}
    \toprule
    Dataset & \multicolumn{2}{c}{CIFAR-10}      & \multicolumn{2}{c}{CIFAR-100} \\
    \midrule
    \multicolumn{1}{c}{Bit-widths} & \multicolumn{1}{c}{SBM} & \textbf{CDT (Proposed)} & \multicolumn{1}{c}{SBM} & \textbf{CDT (Proposed)}  \\
    \midrule
    4      & 90.91   & \textbf{91.45 (+0.54)}   & 63.82  & \textbf{64.18 (+0.36)}  \\
    8      & 92.78  & \textbf{93.03 (+0.25)}    & 66.71  & \textbf{67.45 (+0.74)}  \\
    12     & 92.75  & \textbf{93.06 (+0.31)}    & 67.13  & \textbf{67.42 (+0.29)}  \\
    16     & 92.90   & \textbf{93.09 (+0.19)}    & 67.17  & \textbf{67.50 (+0.33)} \\
    32     & 92.5 \ & \textbf{93.08 (+0.58)}    & 67.18  & \textbf{67.47 (+0.29)}  \\
    \midrule
    \midrule
    4      & 90.91   & \textbf{91.88 (+0.97)}   & 63.82  & \textbf{64.12 (+0.30)} \\
    5      & 92.35  & \textbf{92.56 (+0.21)}    & 66.20   & \textbf{66.68 (+0.48)} \\
    6      & 92.80   & \textbf{92.93 (+0.13)}    & 66.48  & \textbf{66.55 (+0.07)} \\
    8      & 92.78  & \textbf{93.02 (+0.24)}    & 66.71  & \textbf{66.88 (+0.17)} \\
    \bottomrule
    \end{tabular}%
    }
  \label{tab:resnet38}%
  \vspace{-5pt}
\end{table}%

\begin{table}[!t]
  \centering
  \caption{CDT over independently trained SBM~\cite{banner2018scalable} on \textbf{ResNet-74}, where the values in the bracket represent CDT's accuracy gain over SBM (the higher, the better). }
  \resizebox{0.8\linewidth}{!}{
    \begin{tabular}{cccccccc}
    \toprule
    Dataset &\multicolumn{2}{c}{CIFAR-10}      & \multicolumn{2}{c}{CIFAR-100} \\
    \midrule
    \multicolumn{1}{c}{Bit-widths} & \multicolumn{1}{c}{SBM} & \textbf{CDT (Proposed)} & \multicolumn{1}{c}{SBM} & \textbf{CDT (Proposed)}  \\
    \midrule
    4      & 91.82  & \textbf{92.34 (+0.52)}  & 66.31  & \textbf{67.35 (+1.04)}  \\
    8      & 93.22  & \textbf{93.56 (+0.34)}  & 69.85  & \textbf{69.98 (+0.13)}  \\
    12     & 93.26  & \textbf{93.53 (+0.27)}  & 69.97  & \textbf{69.99 (+0.02)}  \\
    16     & 93.40  & \textbf{93.51 (+0.11)}  & 69.92  & \textbf{70.01 (+0.09)}  \\
    32     & 93.38  & \textbf{93.49 (+0.11)}  & 69.46  & \textbf{69.98 (+0.52)}  \\
    \midrule
    \midrule
    4      & 91.82  & \textbf{92.51 (+0.69)}  & 66.31  & \textbf{67.34 (+1.03)} \\
    5      & 92.98  & \textbf{93.54 (+0.56)}  & 68.66  & \textbf{69.49 (+0.83)} \\
    6      & 93.19  & \textbf{93.47 (+0.28)}  & 69.42  & \textbf{69.65 (+0.23)} \\
    8      & 93.22  & \textbf{93.72 (+0.50)}  & 69.85  & \textbf{70.02 (+0.17)} \\
    \bottomrule
    \end{tabular}%
    }
  \label{tab:resnet74}%
  \vspace{-1.5em}
\end{table}%

\begin{table}[btp]
    \vspace{-1.5em}
    \centering
    \caption{CDT over SP~\cite{banner2018scalable} on ResNet-18 and TinyImageNet in terms of test accuracy, where the values in the bracket represent CDT's accuracy gain over SBM. }
    \begin{tabular}{cc|cc}
    \toprule
    \multicolumn{2}{c}{Bit-widths} & \multicolumn{2}{c}{Methods} \\
    \midrule
    Weight & Activation & SP & \textbf{CDT (Proposed)} \\
    \midrule
    2      & 2    &  47.8  &  \textbf{52.3 (+4.5)}  \\
    2      & 32   &  50.5  &  \textbf{51.3 (+0.8)}  \\
    32     & 2    &  51.8  &  \textbf{53.4 (+1.6)}  \\
    \bottomrule
    \end{tabular}
    \label{tab:tinyimagenet}
\end{table}

We also benchmark CDT on ResNet-38/74~\cite{wang2018skipnet} with CIFAR-10/CIFAR-100 over independently trained SBM~\cite{banner2018scalable}. As shown in Tab.~\ref{tab:resnet38} and~Tab.~\ref{tab:resnet74} for ResNet-38 and ResNet-74, respectively, CDT consistently achieves a better/comparable accuracy (0.02\%$\sim$1.04\%) over the independently trained ones under all the models/datasets/bit-widths, and notably boosts the accuracy of the lowest bit-width (4-bit) by 0.30\%$\sim$1.04\%.

To evaluate CDT's performance when involving extremely low bit-width (2-bit), we further benchmark CDT on ResNet-18~\cite{he2016deep} and TinyImageNet~\cite{le2015tiny} over the SP~\cite{banner2018scalable} baseline. The results are shown in Tab.~\ref{tab:tinyimagenet}. It can be observed that the CDT is particularly effective in boosting the accuracy in lower bit-widths.
Specifically, when the weights and activations both adopt 2-bit, the proposed CDT achieves a 4.5\% higher accuracy than that of the baseline SP method.


\begin{figure}[!tb]
    \vspace{-0.3em}
    \centering
    \includegraphics[width=0.45\textwidth]{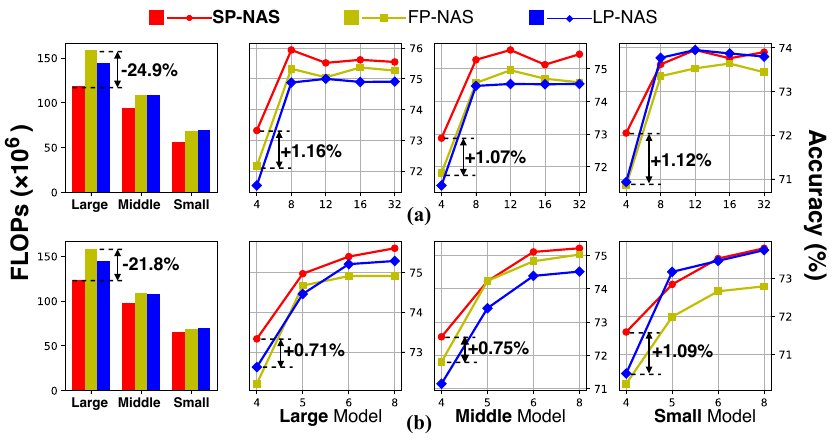}
        \vspace{-0.2cm}
    \caption{InstantNet's SP-NAS over Full-Precision-NAS (FP-NAS) and Low-Precision-NAS (LP-NAS) on CIFAR-100 under large, middle, and small FLOPs constraints trained for two bit-width sets: (a) [4, 8, 12, 16, 32], and (b) [4, 5, 6, 8].}
    \label{fig:exp_spnas}
\end{figure}

\vspace{-0.1cm}
\subsection{Ablation study of InstantNet: SP-NAS}
\label{sec:exp_nas}
\vspace{-0.1cm}



From Fig.~\ref{fig:exp_spnas}, we can see that: (1) SP-NAS consistently outperforms the baselines at the lowest bit-width, which is the bottleneck in SOTA SP-Nets~\cite{jin2019adabits}, while offering a higher/comparable accuracy at higher bit-widths. Specifically, SP-NAS achieves a 0.71\%$\sim$1.16\% higher accuracy over the strongest baseline at the lowest bit-width on both bit-width sets under the three FLOPs constraints; and (2) SP-NAS shows a notable superiority on the bit-width set with a larger dynamic range which is more favorable for IoT applications as larger bit-width dynamic ranges provide more flexible instantaneous accuracy-efficiency trade-offs. Specifically, compared with the strongest baseline, SP-NAS achieves a 1.16\% higher accuracy at the lowest bit-width and a 0.25\%$\sim$0.61\% higher accuracy at other bit-widths, while offering a 24.9\% reduction in FLOPs on the bit-width set [4, 8, 12, 16, 32]. This experiment validates that SP-NAS can indeed effectively tackle SP-Nets' bottleneck and improve its scalability over previous search methods which fail to guarantee accuracy at lower bit-widths.


\begin{figure}[!tb]
    \vspace{-1em}
    \centering
    \includegraphics[width=0.4\textwidth]{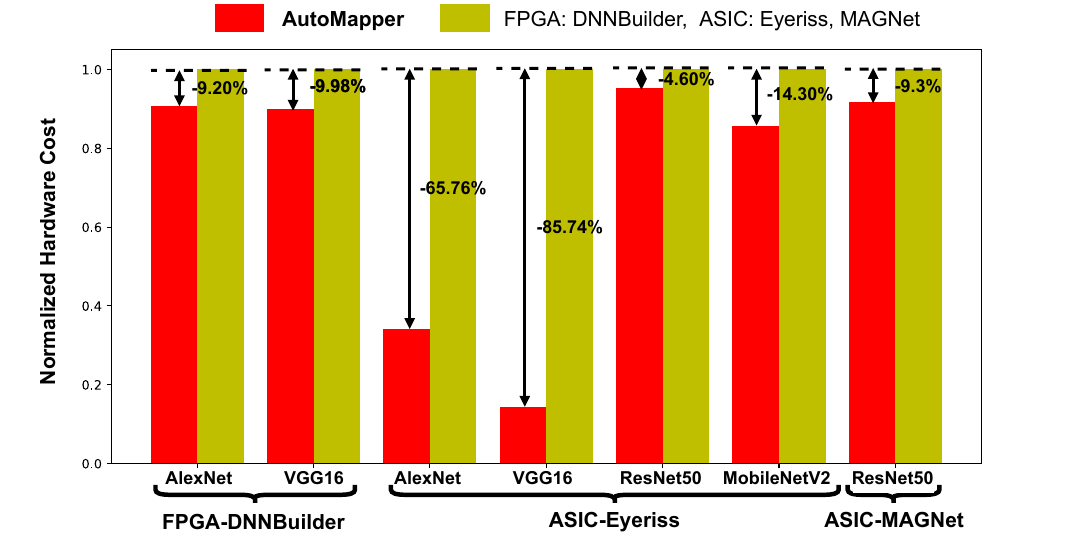}
    \caption{AutoMapper over SOTA expert-crafted and tool generated dataflows on FPGA/ASIC.}
    \label{fig:exp_automapper}
    \vspace{-1.5em}
\end{figure}

\subsection{Ablation study of InstantNet: AutoMapper}
\label{sec:exp_mapping}
As shown in Fig.~\ref{fig:exp_automapper}, we can see that (1) the dataflows suggested by AutoMapper (taking less than 10 minutes of search time) even outperforms SOTA expert-crafted designs: the mapping generated by AutoMapper achieves 65.76\% and 85.74\% EDP reduction on AlexNet~\cite{krizhevsky2012imagenet} and VGG16~\cite{simonyan2014very} compared with Eyeriss~\cite{eyeriss}, respectively; (2) AutoMapper achieves a higher cost savings on ASIC than that of FPGA. This is because ASIC designs are more flexible than FPGA in their dataflows and thus achieve superior performance when exploring using effective automated search tools; and (3) when comparing with MAGNet, we have roughly 9.3\% reduction in terms of the energy cost. MAGNet only used a pre-defined set of loop-orders to cover different dataflow scenarios, which may not generically fit network's diverse layer structures, thus resulting in inferior performance. 

\begin{figure}[!tb]
    \vspace{-0.5em}
    \centering
    \includegraphics[width=0.5\textwidth]{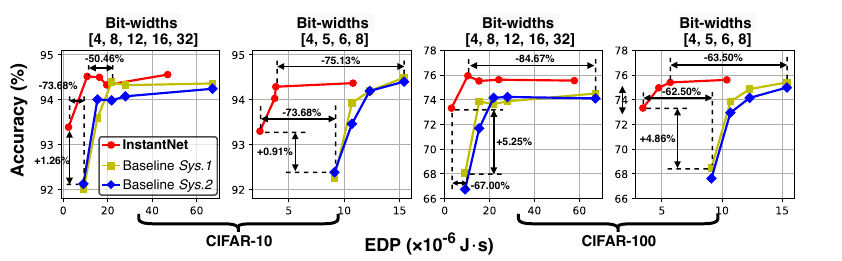}
        \vspace{-2em}
    \caption{InstantNet generated and SOTA IoT systems on CIFAR-10/100 under two bit-width sets. 
     }
 \label{fig:exp_final_cifar}
    \vspace{-1.8em}
\end{figure}

 \vspace{-0.2em}
\subsection{InstantNet over SOTA systems}
\label{sec:exp_sota}

\begin{wrapfigure}{r}{0.25\textwidth}
\vspace{-2em}
  \begin{center}
    \includegraphics[width=0.25\textwidth]{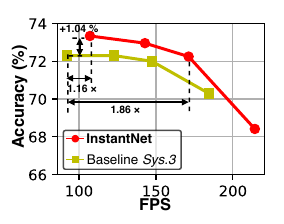}
  \end{center}
  \vspace{-1.5em}
    \caption{InstantNet and SOTA IoT systems on ImageNet with bit-widths of $[4, 5, 6, 8]$.}
    \label{fig:exp_final_imagenet}
    \vspace{-2em}
\end{wrapfigure}

\textbf{Results and analysis on CIFAR-10/100.}
As shown in Fig.~\ref{fig:exp_final_cifar}, we can see that (1) InstantNet generated systems consistently outperforms the SOTA baselines in terms of the trade-off between accuracy and EDP (a commonly-used hardware metric for ASIC) by achieving a higher or comparable accuracy and better EDP under lower bit-widths over the baselines. In particular, InstantNet can achieve up to 84.67\% reduction in EDP with a 1.44\% higher accuracy on CIFAR-100 and the bit-width set of $[4, 8, 12, 16, 32]$;
and (2) InstantNet always surpasses the SOTA baselines under the bottleneck bit-width, i.e., the lowest one, with a 62.5\%$\sim$73.68\% reduction in EDP and a 0.91\%$\sim$5.25\% higher accuracy, which is notably more practical for real-world IoT deployment.


\textbf{Results and analysis on ImageNet.} As shown in Fig.~\ref{fig:exp_final_imagenet}, InstantNet generated system achieves a $1.86\times$ improvement in Frame-Per-Second (FPS) while having a comparable accuracy (-0.05\%) over the SOTA FPGA based IoT system. 
\vspace{-0.3em}
\section{Conclusion}
We propose an \textit{automated} framework termed \textbf{InstantNet} to automatically search for SP-Nets (i.e., capable of operating at variable bit-widths) that can achieve the same or even better accuracy than DNNs optimized for individual bit-widths, and to generate optimal dataflows to maximize efficiency when DNNs are executed under various bit-widths on different devices. 
Extensive experiments show that 
InstantNet has promised an effective automated framework for expediting development and deployment of efficient DNNs for numerous IoT applications with diverse specifications.

\section*{Acknowledgement}
The work is supported by the National Science Foundation (NSF) through the Energy, Power, Control, and Networks (EPCN) program (Award number: 1934755, 1934767).

\bibliographystyle{IEEEtran}
\bibliography{ref}

\end{document}



\title{Supplementary Material \\ InstantNet: Automated Generation and Deployment of Instantaneously Switchable-Precision Networks}
\newcommand{\FrameworkName}{TODO}
\newcommand{\eg}{e.g.,}
\newcommand{\ie}{i.e.,}
\definecolor{Note_color}{rgb}{1.0, 0.0, 0.0}
\newcommand{\NOTE}[1]{\textcolor{Note_color}{#1}}

\vskip 0.3in

]

\section{More ablation studies of Cascade Distillation Training (CDT)}
\subsection{Visualization of prediction distribution}
\begin{figure*}[!bt]
    \centering
    \includegraphics[width=\textwidth]{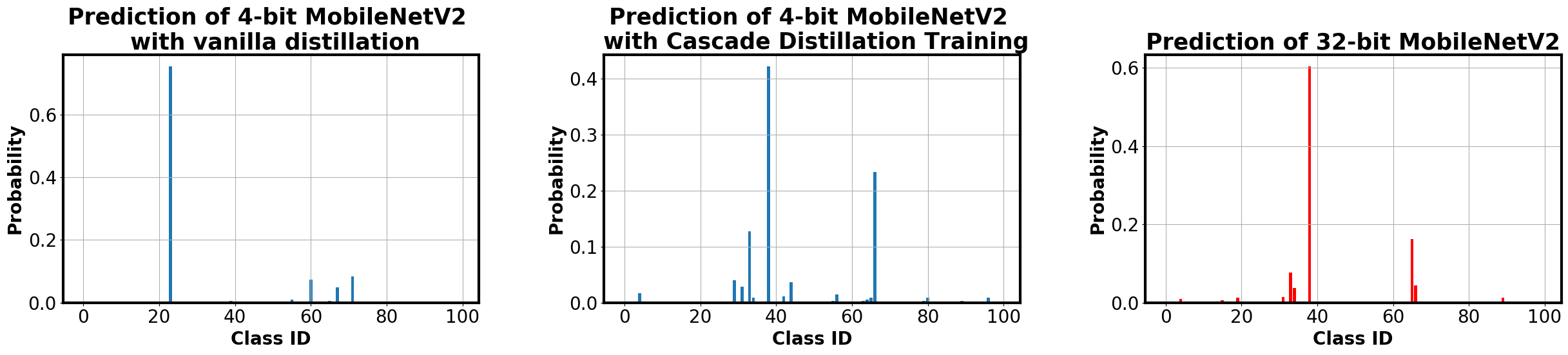}
    \caption{Prediction distribution of MobileNetV2 on CIFAR-100 under \textbf{(left)}: 4-bit trained by vanilla distillation, \textbf{(middle)} 4-bit trained by the proposed Cascade Distillation Training and \textbf{(right)} 32-bit trained by vanilla distillation.}
    \label{fig:output}
\end{figure*}

We visualize the prediction distribution (classification probability after softmax) of MobileNetV2 on CIFAR-100 under the bit-width set of 4, 8, 12, 16, 32 (quantized by SBM~\cite{banner2018scalable}) trained by different strategies in Fig.~\ref{fig:output}. 
In particular, we show prediction distribution of the following three cases with a random sampled image from the test dataset in Fig.~\ref{fig:output} to verify the effectiveness of CDT: (1) 4-bit trained by vanilla distillation, i.e., only consider the distillation with 32-bit width, (2) 4-bit trained by the proposed CDT technique and (3) 32-bit trained by vanilla distillation. We can observe that vanilla distillation fails to narrow the gap between 32-bit and the lowest 4-bit due to the large quantization error. This is actually a common phenomenon among the test dataset based on the observation that the validation accuracy of the 4-bit network is around 1\%, denoting the failure of vanilla distillation for tackling the bit-width set with large dynamic range. While the proposed CDT notably helps the prediction distribution of the 4-bit network smoothly evolve to that of the 32-bit one, and also boost its validation accuracy to 71.21\%, verifying the superiority of CDT.

\subsection{Experiments on different networks}
To further verify the general effectiveness of the proposed CDT in addition to Sec. 4.2 in the main content, we benchmark CDT on ResNet-38/74~\cite{wang2018skipnet} with CIFAR-10/CIFAR-100 and bit-width sets of $[4,8,12,16,32]$ / $[4,5,6,8]$ over the independently trained SBM~\cite{banner2018scalable} baseline. For consistency and fair comparison, we apply the same training setting as in Sec. 4.2. As shown in Tab.~\ref{tab:resnet38} and~\ref{tab:resnet74} for ResNet-38 and ResNet-74 respectively, we can observer that (1) CDT consistently achieves better or comparable accuracy (0.02\%$\sim$1.04\%) compared with the independently trained ones under all the models, datasets and bit-width settings. (2) CDT notably boosts the accuracy of the lowest bit-width (4-bit) by 0.30\%$\sim$1.04\%, further proving CDT's capability of tackling the bottleneck bit-width in SP-Nets. 

\begin{table}[htbp]
  \centering
  \caption{Benchmark the proposed CDT over independently trained SBM~\cite{banner2018scalable} on ResNet-38.}
  \resizebox{\linewidth}{!}{
    \begin{tabular}{cccccccc}
    \toprule
    \multicolumn{4}{c}{CIFAR-10}      & \multicolumn{4}{c}{CIFAR-100} \\
    \midrule
    \multicolumn{1}{c}{Bit-widths} & \multicolumn{1}{c}{SBM} & \textbf{CDT (Proposed)} & \multicolumn{1}{c}{\textbf{Improvement}} & \multicolumn{1}{c}{Bit-widths} & \multicolumn{1}{c}{SBM} & \textbf{CDT (Proposed)} & \multicolumn{1}{c}{\textbf{Improvement}} \\
    \midrule
    4      & 90.91  & \textbf{91.45} & \textbf{0.54} & 4      & 63.82  & \textbf{64.18} & \textbf{0.36} \\
    8      & 92.78  & \textbf{93.03} & \textbf{0.25} & 8      & 66.71  & \textbf{67.45} & \textbf{0.74} \\
    12     & 92.75  & \textbf{93.06} & \textbf{0.31} & 12     & 67.13  & \textbf{67.42} & \textbf{0.29} \\
    16     & 92.9   & \textbf{93.09} & \textbf{0.19} & 16     & 67.17  & \textbf{67.5} & \textbf{0.33} \\
    32     & 92.5   & \textbf{93.08} & \textbf{0.58} & 32     & 67.18  & \textbf{67.47} & \textbf{0.29} \\
    \midrule
    \midrule
    4      & 90.91  & \textbf{91.88} & \textbf{0.97} & 4      & 63.82  & \textbf{64.12} & \textbf{0.30} \\
    5      & 92.35  & \textbf{92.56} & \textbf{0.21} & 5      & 66.2   & \textbf{66.68} & \textbf{0.48} \\
    6      & 92.8   & \textbf{92.93} & \textbf{0.13} & 6      & 66.48  & \textbf{66.55} & \textbf{0.07} \\
    8      & 92.78  & \textbf{93.02} & \textbf{0.24} & 8      & 66.71  & \textbf{66.88} & \textbf{0.17} \\
    \bottomrule
    \end{tabular}%
    }
  \label{tab:resnet38}%
\end{table}%

\begin{table}[htbp]
  \centering
  \caption{Benchmark the proposed CDT over independently trained SBM~\cite{banner2018scalable} on ResNet-74.}
  \resizebox{\linewidth}{!}{
    \begin{tabular}{cccccccc}
    \toprule
    \multicolumn{4}{c}{CIFAR-10}      & \multicolumn{4}{c}{CIFAR-100} \\
    \midrule
    \multicolumn{1}{c}{Bit-widths} & \multicolumn{1}{c}{SBM} & \textbf{CDT (Proposed)} & \multicolumn{1}{c}{\textbf{Improvement}} & \multicolumn{1}{c}{Bit-widths} & \multicolumn{1}{c}{SBM} & \textbf{CDT (Proposed)} & \multicolumn{1}{c}{\textbf{Improvement}} \\
    \midrule
    4      & 91.82  & \textbf{92.34} & \textbf{0.52} & 4      & 66.31  & \textbf{67.35} & \textbf{1.04} \\
    8      & 93.22  & \textbf{93.56} & \textbf{0.34} & 8      & 69.85  & \textbf{69.98} & \textbf{0.13} \\
    12     & 93.26  & \textbf{93.53} & \textbf{0.27} & 12     & 69.97  & \textbf{69.99} & \textbf{0.02} \\
    16     & 93.4   & \textbf{93.51} & \textbf{0.11} & 16     & 69.92  & \textbf{70.01} & \textbf{0.09} \\
    32     & 93.38  & \textbf{93.49} & \textbf{0.11} & 32     & 69.46  & \textbf{69.98} & \textbf{0.52} \\
    \midrule
    \midrule
    4      & 91.82  & \textbf{92.51} & \textbf{0.69} & 4      & 66.31  & \textbf{67.34} & \textbf{1.03} \\
    5      & 92.98  & \textbf{93.54} & \textbf{0.56} & 5      & 68.66  & \textbf{69.49} & \textbf{0.83} \\
    6      & 93.19  & \textbf{93.47} & \textbf{0.28} & 6      & 69.42  & \textbf{69.65} & \textbf{0.23} \\
    8      & 93.22  & \textbf{93.72} & \textbf{0.50} & 8      & 69.85  & \textbf{70.02} & \textbf{0.17} \\
    \bottomrule
    \end{tabular}%
    }
  \label{tab:resnet74}%
\end{table}%

\section{Details about AutoMapper}
\subsection{The \textit{for-loop} based representation adopted in the generic design space}




\begin{figure}[!b]
\begin{minipage}{\linewidth}
    \begin{algorithm}[H]
        \label{alg:2dconv}
        \caption{CONV as \textit{for-loop} structure}
        \label{alg:2dconv}
        \begin{algorithmic}[1]
        \vspace{5pt}  
        \STATE{ \textit{for} $s=0: S-1$ \textbf{kernel\_col}}
        \vspace{10pt}
        \STATE{\hspace{8pt}  \textit{for} $r=0: R-1$ \textbf{kernel\_row}}
        \vspace{10pt}
        \STATE{\hspace{16pt}  \textit{for} $x=0: X-1$ \textbf{output\_col}}
        \vspace{10pt}
        \STATE{\hspace{24pt}  \textit{for} $y=0: Y-1$ \textbf{output\_row}}
        \vspace{10pt}
        \STATE{\hspace{32pt}  \textit{for} $k=0: K-1$ \textbf{channel\_out}}
        \vspace{10pt}
        \STATE{ \hspace{40pt}\textit{for} $c=0: C-1$  \textbf{channel\_in}}
        \vspace{10pt}
         \STATE{\hspace{48pt}\color{blue} // MAC operation}
        \STATE{\hspace{48pt}  ofmap[k][y][x]+=
         \\ \vspace{-1em}\hspace{48pt} ifmap[c][y+r][x+s]*kernel[c][k][r][s]}

        \end{algorithmic}
    \end{algorithm}
\end{minipage}
\hfill
\begin{minipage}{\linewidth}
    \begin{algorithm}[H]
        \label{alg:2dconv2}
        \caption{CONV as \textit{for-loop} structure, with \textit{parallel-for}s and \textit{inner-loops} included }
        \label{alg:2dconv2}
        \begin{algorithmic}[1]
        \STATE{\hspace{0pt}  {{\color{blue}// DRAM level }} }
        \vspace{-2pt}
        \STATE{... }
        \vspace{-2pt}
        \STATE{\hspace{0pt}  {{\color{blue}// Global buffer level }} }
        \STATE{\hspace{0pt}  \textit{for} $s_2=0: S_2-1$ \textbf{kernel\_col}}
        \vspace{-2pt}
        \STATE{\hspace{8pt}...}
        \vspace{-2pt}
        \STATE{\hspace{16pt}\textit{for} $c_2=0: C_2-1$  \textbf{channel\_in}}
        \vspace{4pt}
        \STATE{\hspace{0pt}  {{\color{blue}// NoC level}} }
        \STATE{ \textit{parallel-for} $y_1=0: Y_1-1$  \textbf{output\_row}}
        \STATE{ \textit{parallel-for} $x_1=0: X_1-1$  \textbf{output\_col}}
        \vspace{4pt}
        
        \STATE{\hspace{0pt}  {{\color{blue}// RF level}} }
        \STATE{ \textit{for} $c_0=0: C_0-1$  \textbf{channel\_in}}
        \STATE{\hspace{8pt}...}

        \STATE{\hspace{16pt}  \textit{for} $s_0=0: S_0-1$ \textbf{kernel\_col}}
        \STATE{\hspace{24pt}  MAC operation}

        \end{algorithmic}
    \end{algorithm}
\end{minipage}
\end{figure}

For describing DNNs' mapping strategies in real accelerators, one of the most common methods is to implement the \textit{for-loop} description in Alg.\ref{alg:2dconv2} due to its compatibility with standard DNN \textit{for-loop} description (see Alg.\ref{alg:2dconv}) and straightforward representation. Specifically, the representation in Alg.\ref{alg:2dconv2} incorporate the additional primitives: 1) \textit{parallel-for} and 2) \textit{memory-hierarchy} for the base loop structure to cover parallelism and inter-memory-hierarchy data movement in the implemented accelerators. Therefore, our mapping space adopts the same convention for its great generality. The additional primitives will be elaborated below:


\underline{\textit{parallel-for}}: the parallel computation over certain data dimensions. For instance in Alg.\ref{alg:2dconv2}, computation on different rows and columns of output are distributed to $X1*Y1$ PEs and done in parallel.


\underline{\textit{memory-hierarchy}}: Multi-level memory hierarchies are commonly adopted by SOTA accelerators to maximize data reuse and time-energy efficiency. Motivated by this advantage, the description in Alg.\ref{alg:2dconv2} also uses multiple levels of nested loops each of which corresponds to one memory level. For instance, as shown in Alg.\ref{alg:2dconv2}, we extend the original loop structure to multiple levels of loops with each level representing temporal data computation and transfer for intermediate results within each memory, e.g., extending $C$ to $C_0$ and $C_2$ for tiling data between Register File(RF) and Global buffer.
\begin{table}[t]
\centering
\caption{\textbf{Detailed illustration of design factors across different memory hierarchy, with $n$ denoting the number of loops at one single memory hierarchy which also equals to the number of data dimensions.}}
\resizebox{\linewidth}{!}{
\renewcommand\arraystretch{1.2}
\begin{tabular}{c|c|c}
\cline{1-3}
\multicolumn{1}{l|}{\textbf{}}  & Design factors & Search Space \\ \cline{1-3}

\multirow{2}{*}{DRAM} & \textit{loop-order}                       &   $\{ (l_1,..l_n) | \forall i,j, \, i\neq j \rightarrow l_i\neq l_j \}$ \\ \cline{2-3}& \textit{loop-size}  &  $\{ (ls_1,..ls_n) \}$  \\ \cline{1-3} 

\multirow{2}{*}{Global Buffer} & \textit{loop-order} & $\{ (l_1,..l_n) | \forall i,j, \, i\neq j \rightarrow l_i\neq l_j \}$  \\ \cline{2-3}& \textit{loop-size} & $\{ (ls_1,..ls_n)  \}$ \\ \cline{1-3}

\multirow{2}{*}{RF} & \textit{loop-order} & $\{ (l_1,..l_n) | \forall i,j, \, i\neq j \rightarrow l_i\neq l_j \}$  \\ \cline{2-3}& \textit{loop-size} & $\{ (ls_1,..ls_n) \}$ \\ \cline{1-3}

 PE& \textit{Pipeline / Multi-cycle} & $  \{0, 1\} $        \\ \cline{1-3}
\end{tabular}
}
\label{tab: design_factors}
\end{table}

\subsection{Design factors}
As shown in Tab.\ref{tab: design_factors}, the proposed generic design space contains three major design factors:  1) \textit{loop-order}, 2) \textit{loop-size}, and 3) \textit{pipeline/multi-cycle}. Each level of memory and PE array will be paired with a independent set of these design factors as illustrated in Tab.\ref{tab: design_factors}. More details about each design factor are provided below:

\begin{figure}[tbh]
    \vspace{-1.5em}
    \begin{minipage}{\linewidth}
    \begin{algorithm}[H]
    \DontPrintSemicolon
    \KwIn{Efficiency Goal, DNN, Design Space (DS)}
    \KwOut{Optimal algorithm-to-device mapping}

      Build a $pool$ with $n$ random samples from DS
      
      \While{\textit{Efficiency Goal not met}}
      {
        \uIf{$size(pool) \leq n$}
        {
          \For{$m$ iterations}
          {

            Random Pick  $p  \in pool$\;
            Random Perturb $k$ features of $p$

          Add $p$ to $pool$
          
         }
        }
        \Else
        {
            Rank the samples in $pool$\ with given DNN
            
            Remove worst $m$ samples from  \textit{pool}\; 
        }
      }
      \Return{optimal mapping in $pool$}
    \caption{Evolutionary AutoMapper}\label{alg:eaalg}
    \end{algorithm}
\end{minipage}
\vspace{-2em}
\end{figure}


\textbf{\textit{loop-order}}: the order of the loops in a specific memory level, which includes $n$ loops, equal to the total number of data dimensions (six in the case of Alg.\ref{alg:2dconv}). Deciding the \textit{loop-order} is then an $n$-item ordering problem, which can be solved by picking a loop, such as loop over input channels, for each of the $n$ positions in the loop structure at each memory level. Then, the searchable choices can be represented with a list of unique integers, as in Tab.\ref{tab: design_factors}, with the element at list position $i$ denoting a certain loop  at position $i$ of the nested loop structure.

\textbf{\textit{loop-size}}:
the size of each loop in the \textit{for-loop} structure, which follows the principle that the product of all \textit{loop-size}s belonging to one data dimension should equal to the size of that specific data dimension in the given DNN structure. For instance in Alg.\ref{alg:2dconv2}, $C_0*C_2*C_3=C$ (input channel size). Therefore, we generate all the possible choices for \textit{loop-size}s from the factorization of each DNN data dimension respectively. The final choice of a set of \textit{loop-size}s will subject to (1) the product should equal to the corresponding data dimension, and (2) memory size constraints. The final choices are in form of a list of integers with size $n$.

\textbf{\textit{Pipeline/multi-cycle}}: a binary choice between a layer-wise pipeline hardware architecture and a multi-cycle hardware architecture shared by all layers.

\subsection{Detailed evolutionary search algorithm}

To deal with the prohibitively large mapping space, we adopt an evolutionary based search algorithm, considering that evolutionary algorithms have more exploitation than random search while better suiting the highly discrete space \cite{google_ev,genesys}.

We summarize our Evolutionary AutoMapper in Alg.\ref{alg:eaalg}. By keeping filtering out the worst performing samples and perturbing the well performing samples, the AutoMapper balances the exploration of the generic design space and the exploitation of searched advantageous mappings. In particular, the population size $n$ is set to be $1000$ and $m$ and $k$, the number of samples and features to be updated, are set to be 30\% of the population size and 30\% of the total number of features respectively.

\section{Details about experiment setup}
\subsection{Hardware setup}

\underline{Mapping implementation methodology:} 
we consider two commonly used IoT hardware platforms, i.e., ASIC and FPGA, for evaluating our AutoMapper. Specifically, for FPGA setup, we choose the  Vivado HLx design tool-flow where we first synthesize the accelerator design in c++ via Vivado HLS, and then plug the HLS exported IPs into a Vivado IP integrator to generate the corresponding bit streams, which are programmed into the FPGA board for on-board executio and measurements; for ASIC setup, we synthesize Verilog designs, from the generated mapping, with a Synopsys Design Compiler, based on a commercial CMOS technology, and then place and route in a Synopsys IC Compiler for obtaining its actual area.


\subsection{Software setup}
\textbf{Search and training on CIFAR-10/100.} \underline{Search space.} We adopt the same search space as~\cite{wu2019fbnet} except the stride settings for each group (i.e., a set of blocks with the same number pf output channels) to adapt to the input resolution of CIFAR-10/100. We follow the stride settings as MobileNetV2 on CIFAR-10/100 in~\cite{wang2019e2}, i.e. $[1, 1, 2, 2, 1, 2, 1]$ for the seven groups.
\underline{Search settings.} We search for the optimal DNN network in 50 epochs with a batch size of 64. In particular, we update the supernet weights with cascade distillation as formulated in Eq.(2) on half of the training dataset with an SGD optimizer, a momentum of 0.9, and an initial learning rate of 0.025 with cosine decay, and update network architecture parameters with the lowest bit-width as in Eq.(2) on the other half of the training dataset with an Adam optimizer, a momentum of 0.9, and a fixed learning rate of 3e-4. We apply gumbel softmax on the architecture parameters as the contributing coefficients of each option to the supernet (following~\cite{wu2019fbnet}), where the initial temperature is 3 and decays by 0.94 at the end of each epoch.  
\underline{Train from scratch.} For training the derived network architectures from scratch with the proposed Cascade Distillation Training (CDT), we adopt an SGD optimizer with a momentum of 0.9, an initial learning rate of 0.025 with cosine decay for 200 epochs and a batch size of 128. 

\textbf{Search and training on ImageNet.} \underline{Search space.} We adopt the same search space as~\cite{wu2019fbnet}.
\underline{Search settings.} We follow the same hyper-parameter settings for network search as ~\cite{wu2019fbnet} on ImageNet. In particular, with the proposed SP-NAS algorithm in Eq.(2), we search for the optimal network for 90 epochs with a batch size of 192, update the supernet weights on 80\% of the training dataset with an SGD optimizer, a momentum of 0.9, and an initial learning rate of 0.1 with the cosine decay, and update the network architecture parameters on the remaining 20\% training dataset with an Adam optimizer, a momentum of 0.9, and a fixed learning rate of 1e-2. The initial temperature of gumbel softmax is 5 and decays by 0.956 every epoch.
\underline{Train from scratch.} Following~\cite{wu2019fbnet}, we apply the proposed CDT to train the derived network with an SGD optimizer, a momentum of 0.9, an initial learning rate of 0.1 which decays by 0.1 at the 90-th, 180-th, 270-th epoch, respectively, within the total 360 epochs and a batch size of 256.

\bibliography{ref}
\bibliographystyle{IEEEtran}